# Classification of Documents Extracted from Images with Optical Character Recognition Methods

*Ömer AYDIN*[1*]
*[*1]Manisa Celal Bayar Üniversitesi, Mühendislik Fakültesi, Manisa, Türkiye*
*(omer.aydin@cbu.edu.tr, Orcid: 0000-0002-7137-4881)*

*Abstract*— Over the past decade, machine learning methods have given us driverless cars, voice recognition, effective web search, and a much better understanding of the human genome. Machine learning is so common today that it is used dozens of times a day, possibly unknowingly. Trying to teach a machine some processes or some situations can make them predict some results that are difficult to predict by the human brain. These methods also help us do some operations that are often impossible or difficult to do with human activities in a short time. For these reasons, machine learning is so important today. In this study, two different machine learning methods were combined. In order to solve a real-world problem, the manuscript documents were first transferred to the computer and then classified. We used three basic methods to realize the whole process. Handwriting or printed documents have been digitalized by a scanner or digital camera. These documents have been processed with two different Optical Character Recognition (OCR) operation. After that generated texts are classified by using Naive Bayes algorithm. All project was programmed in Microsoft Visual Studio 12 platform on Windows operating system. C# programming language was used for all parts of the study. Also, some prepared codes and DLLs were used.

## 1.Introduction

Machine learning is a method that allows computers to act without programmed. This method has been used in many processes such as driverless vehicle technologies, audio and image processing techniques, face recognition systems, and quality control systems. Machine learning has become so common today that it is used in many areas of daily life without realizing it. This method, which is of great importance for the development of artificial intelligence, is shown by many researchers as the best way to develop human-level artificial intelligence. The machine learning methods can be used in any area in our life to help people to make their life easy. Trying to teach some process or some situations to a machine can make them to estimate some result which are difficult to estimate by human brain. These methods also help us to do some operations which are usually impossible or hard to make in short time with human activities. Because of these reasons, machine learning is so important in recent world.

The texts we call manuscripts are texts that have not been printed on a mechanical printing machine or printed on any automated printing device. These documents are hand-written by humans. Before the printing house was invented, all books and documents were written by hand. These writings could be in the form of book, codex, or parchment.

In this study, two different machine learning methods were combined. In order to solve a real-world problem, the manuscript documents were first transferred to the computer and then classified. We used three basic methods to realize the whole process. Created handwritings and computer-printed documents were used. These documents were transferred back to the digital environment with the help of a camera or scanner. These digital documents, which were re-created in the computer environment, were subjected to two different OCR processes. Classification of the texts extracted by OCR was provided using the Naive Bayes algorithm. All project was programmed in Microsoft Visual Studio 12 platform on Windows operating system. C# programming language was used for all parts of the study. Also, some prepared codes and DLLs were used.

## 2. Literature Review

There are many publications in the literature about the methods used in this study. Studies generally focused on one method. Especially optical character recognition and text mining and classification of documents are the most frequently used methods. On the other hand, there are very few studies in the literature that produce an integrated solution by combining two methods as in our study.

Sun et al. published an article which is also a good source for learning OCR process. The article recommends the Smart OCR post-processing system. Machine learning techniques have been used as a semi-automatic solution to reduce human-machine interaction and correct optical recognition errors. Adaptiveness is one of the most important feature of the proposed system. The proposed system can perform well in any environment and adapt immediately. The article mentions that the strategy is applicable to all languages, but is only used for the English language. (Sun et al., 1992).

Holmes et al. published an article in IEEE at 1994. This is an old article but it was useful to read this article as a starting point for this tool. This tool is very useful for machine learning because you can test or run your data in this tool by using different machine learning methods. This tool is a complete workbench so you can cross check between supervised/unsupervised methods. In this article, you can find a detailed definition of the earlier version of WEKA software (Holmes et al, 1994).

Naive Bayes is known to have satisfactory results in many data mining operations. On the contrary, it cannot produce very successful results in text classification processes. Kim et al. found that Naive Bayes was weak in text classification in natural language texts. The authors proposed two empirical heuristics in their articles. In situations where the training set is inadequate, it can be difficult to create an appropriate probabilistic classification model. They discussed increasing the classification weights for such rare categories. (Kim et al., 2006).

Cord and Cunningham published a book in 2008. In the chapter 2, supervised learning has been described. The main topic is using supervised learning methods for multimedia data. At the beginning, “Statistical Learning” was examined after that “Support Vector Machines”, “Nearest Neighbour Classification” and “Ensemble Techniques” have been detailed in this chapter. It is mentioned that using ensemble techniques usually gives better result from using single methods (Cord and Cunningham, 2007).

Qiang stated that Naive Bayes was not sufficient in terms of performance, on the other hand, he was flawed due to reasons such as not well modeling the text and inappropriate feature selection. Some changes are proposed in this article to improve the flaws of Naive Bayes. In the article, experiments were made with this algorithm in the Spam filter category and the results were compared with previous studies (Qiang, 2010).

In an article written by Singh and Budhiraja, mainly feature extraction methods are defined, k-NN, SVM (Support Vector Machines) and Probabilistic Neural Network (PNN) classifiers were compared and whole process of OCR was explained. It has been declared by the authors that Zoning is used as a feature extraction method. An accuracy of 73.02% was achieved. The accuracy ratio was obtained using SVM (Polynomial Kernel). Using the zoning density, SVN, and background directional distribution properties, 95.04% accuracy is achieved with the RBF core. This value is the highest density achieved (Singh and Budhiraja, 2011).

Manchanda et al. published a study in which they proposed a completely new dimension for web page classification using Artificial Neural Networks (ANN). First of all, it examines Naive Bayes and Decision Tree classifiers and compare them for the web page classification. At the end, the paper analyses Neural Network technique. As a result, with this proposed model, the web page classification technique was introduced for the efficient and fast operation of search engines. In addition, the classification results are expected to have high accuracy (Manchanda et al., 2012).

Wemhoener et al. pointed a new thing to decrease the noisy characters (errors) in the OCR generated texts. It is known that so many books on the internet have different versions that is shared on the internet so in this paper authors pointed that these editions can be useful to decrease the errors after OCR operations. If you run OCR for same document the errors will be the same but if you run it for different editions of same book so errors will differ so you can match the right characters from other versions. The proposed approach uses a fast scheme for the alignment of text pairs. A voting scheme is used to correct OCR errors with the alignment process (Wemhoener et al., 2013).

Dessai and Patil proposed an OCR based on Convolutional Neural Network (CNN) to recognise handwritten Devanagari Script. They trained the Convolutional neural networks by using the labelled datasets. They sampled 15 characters from the handwritten text created by different individuals. The training was performed by using 1500 samples of each character. On the other hand, 250 samples were used for the test. After training operation, unknown characters were tested. Accuracy was approximately %89 and %91 which differs by including and excluding some characters. (Dessai and Patil,2019).

Gan et al. proposed a 1-dimensional convolutional neural network (1-D CNN) for online handwritten Chinese character recognition (OLHCCR). When this method compared with 2 dimensional CNN they claimed that their results are better using a more compact model. They also give empirical results to prove their claims (Gan et al.,2019).

Memon et al. wrote a review paper to give a detailed summary of the research papers. These research articles are about the optical character recognition of the handwritten documents. The papers were published between the years 2000 and 2019. They used keywords and forward references investigation method to make this review paper. 176 articles are included in this paper (Memon et al., 2020). This review article is a good starting point to reach and read the related articles so new technologies used for OCR. Moreover, Minaee et al. created an article reviewing more than 150 paper. These papers are about the learning-based text classification models. Also, more than 40 widely used text classification datasets are summarized in their study. As a result, they give an analysis for the different deep learning models' performance (Minaee et al.,2020).

Alrehali et al. aimed to transform Arabic manuscript texts, which are difficult to read, into a publishable form. The texts in image will be converted into readable and processable format with a digital recognition method. This method consists of 4 steps. First, the image will be pre-processed, then the image lines and characters will be created and this stage is segmentation. Thirdly, Arabic character data will be created and finally, the texts will be extracted by classification. (Alrehali et al., 2020).

## 3. Methods

In this study, three basic methods to realize whole process were used. Handwriting or printed documents have been digitalized by scanner or digital camera. This documents have been processed with two different OCR (supervised learning) operation. After that generated texts are classified by using Naive Bayes algorithm (supervised learning). It will be explained whole process and the algorithms in subsections.

All project was programmed in Microsoft Visual Studio 12 platform on Windows operating system. C# programming language was used for all parts of the project. Also some prepared codes, DLLs were used.

### 3.1 Machine Learning OCR Tool

For this tool, first of all we have to fill a document with all alphabetical characters. You can see these documents in Figure 1. These sheets have 26 columns and 12 rows. Each row was filled by a different character from the English alphabet. Same Columns were filled by the same character. These documents were filled by hand writing.

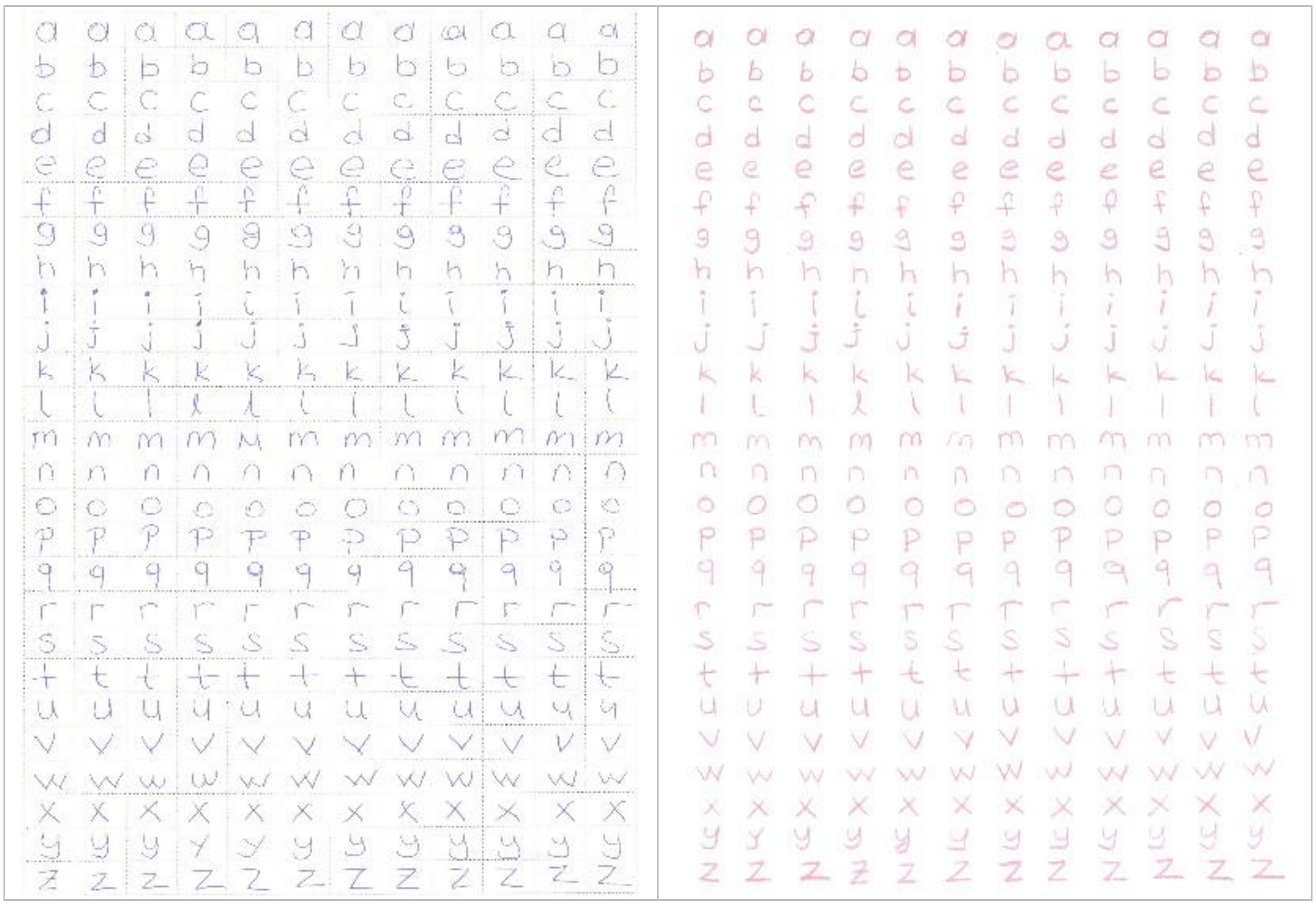

**Figure 1.** Documents filled by handwriting characters.

On the other hand, some sheets were filled it in computer with Arial font and printed shown in Figure 2. After these preparations, these documents were scanned as jpeg format images. After that a text approximately containing 60 words was written about technology. Also a text with same font (Arial) was written using computer and again it was printed. After that this printed document was scanned. So the documents which are filled by English alphabet characters became the training data set. The second ones which are containing the whole text became the test data set for OCR operation.

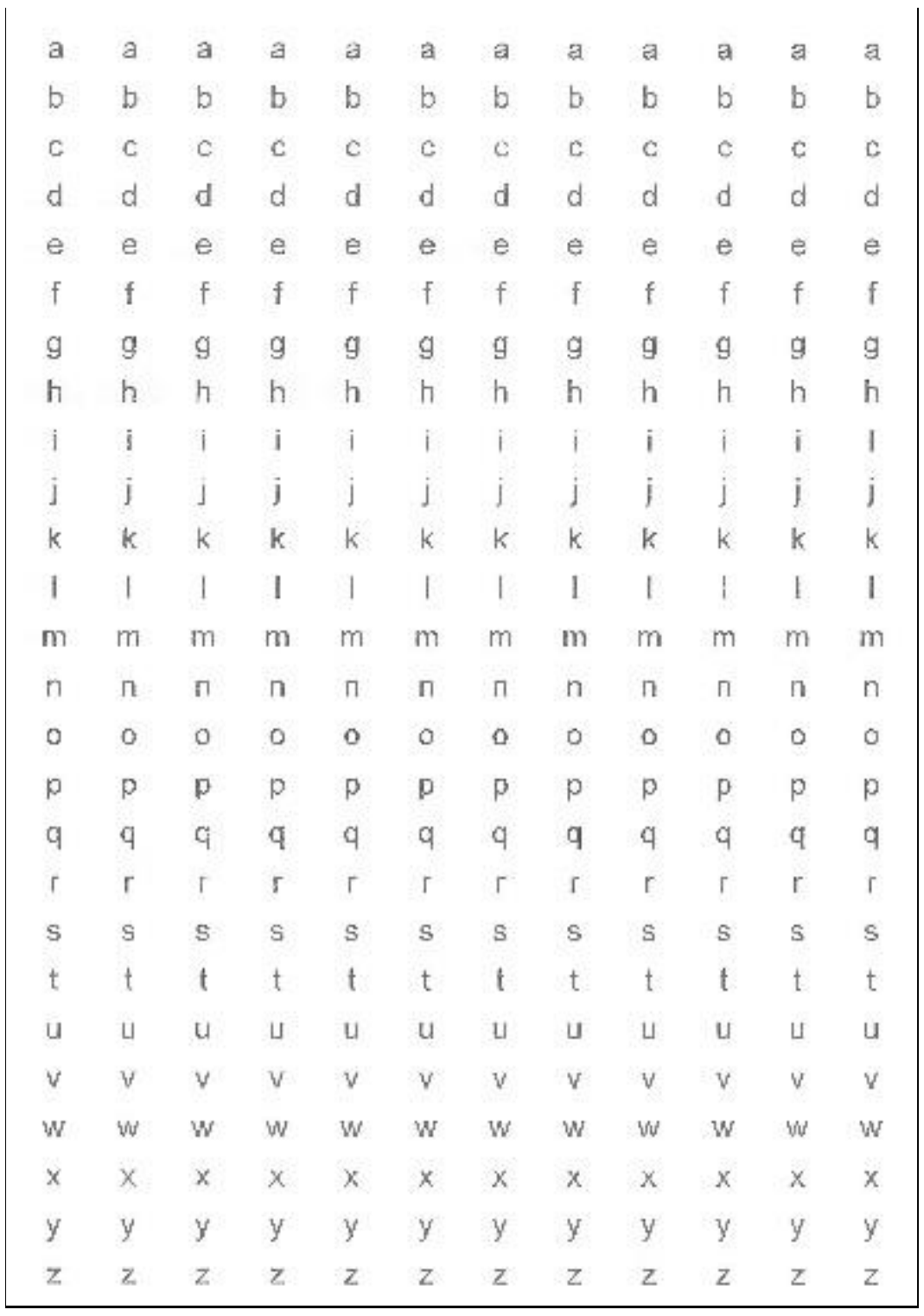

**Figure 2.** Document filled by computer.

First of all, training image which contains the alphabetical characters is opened with the program. We can see this input image in the program's image panel shown in Figure 3. By pressing process button, system automatically crates small blobs for each character. To create these blobs, it creates a grayscale filter, binarized image and invert of the input image. For these processes, system uses AForge framework (Kirillov, 2013).

Small blobs were created which are small image pieces that is containing each character. These blob areas can be changed by manually by user. If the system perceives wrong areas we can change the area by making it smaller, bigger or changing position of it. At the end of this setup process we can now export this small blobs (images) to process with an external tool. Each blob is resized to 20x20 pixel small images. For each blobs system estimate average pixel colour RGB (red, green, blue) values and creates a text file. In this text file for each blob we have three lines. First line is normal image average colour values for each pixel, second line contains binarized image average colour values, the third line contains reversed image average colour values. At the end of each line we can see the number or row that the blob belongs. You can see this text file content in Figure 4.

GNU Octave has capabilities for solving linear and nonlinear problems. It is designed for numerical computations. In addition, it provides data processing and visualization with comprehensive graphic features. MATLAB (Simulink and Natick, 1993) and Octave language are very similar. Thanks to this similarity, it is easy to move codes between MATLAB and Octave (Octave, 2012). The program shown in Figure 5 will be used to estimate them to make a feature extraction method.

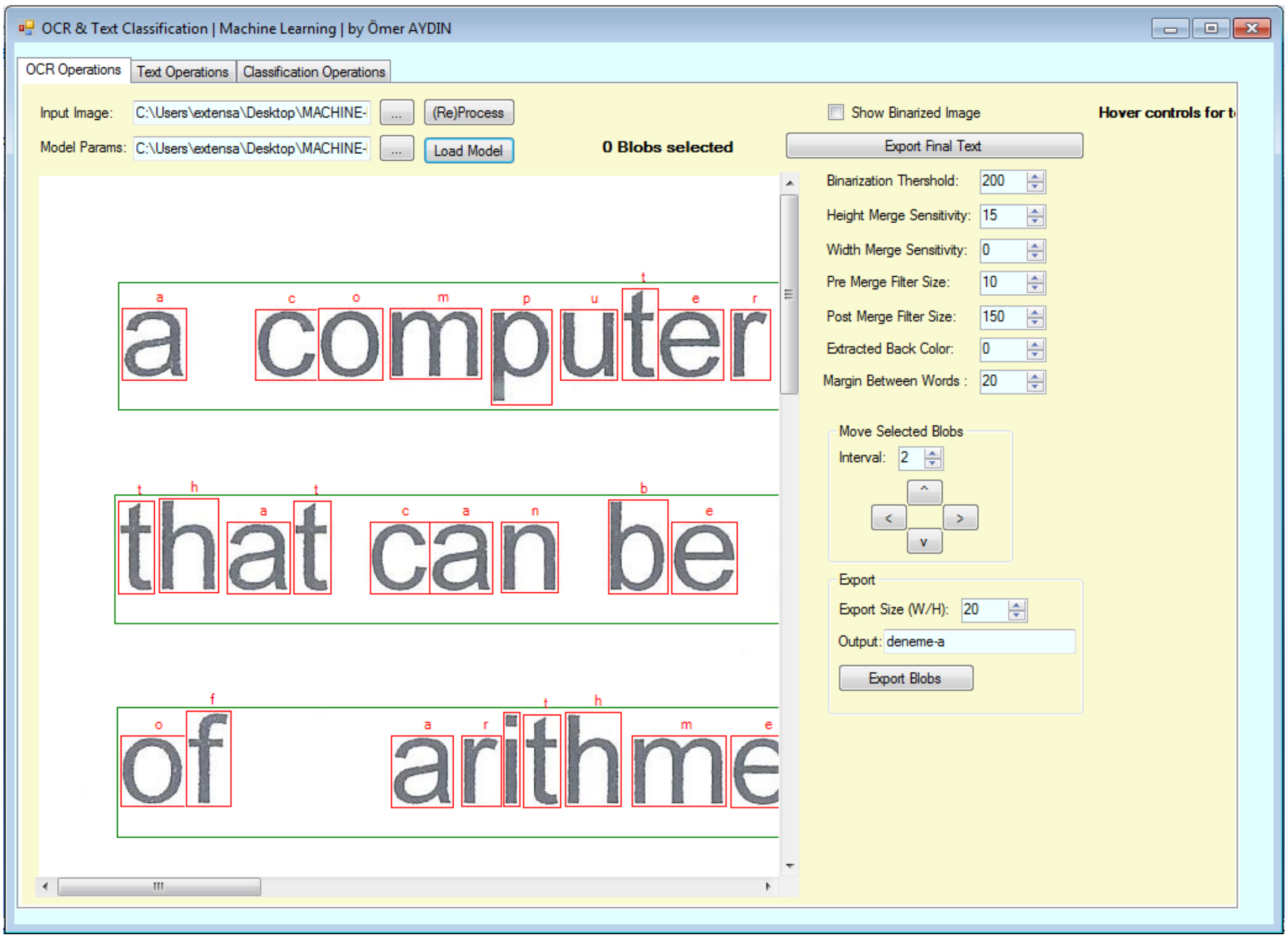

**Figure 3.** OCR Operations program screen.

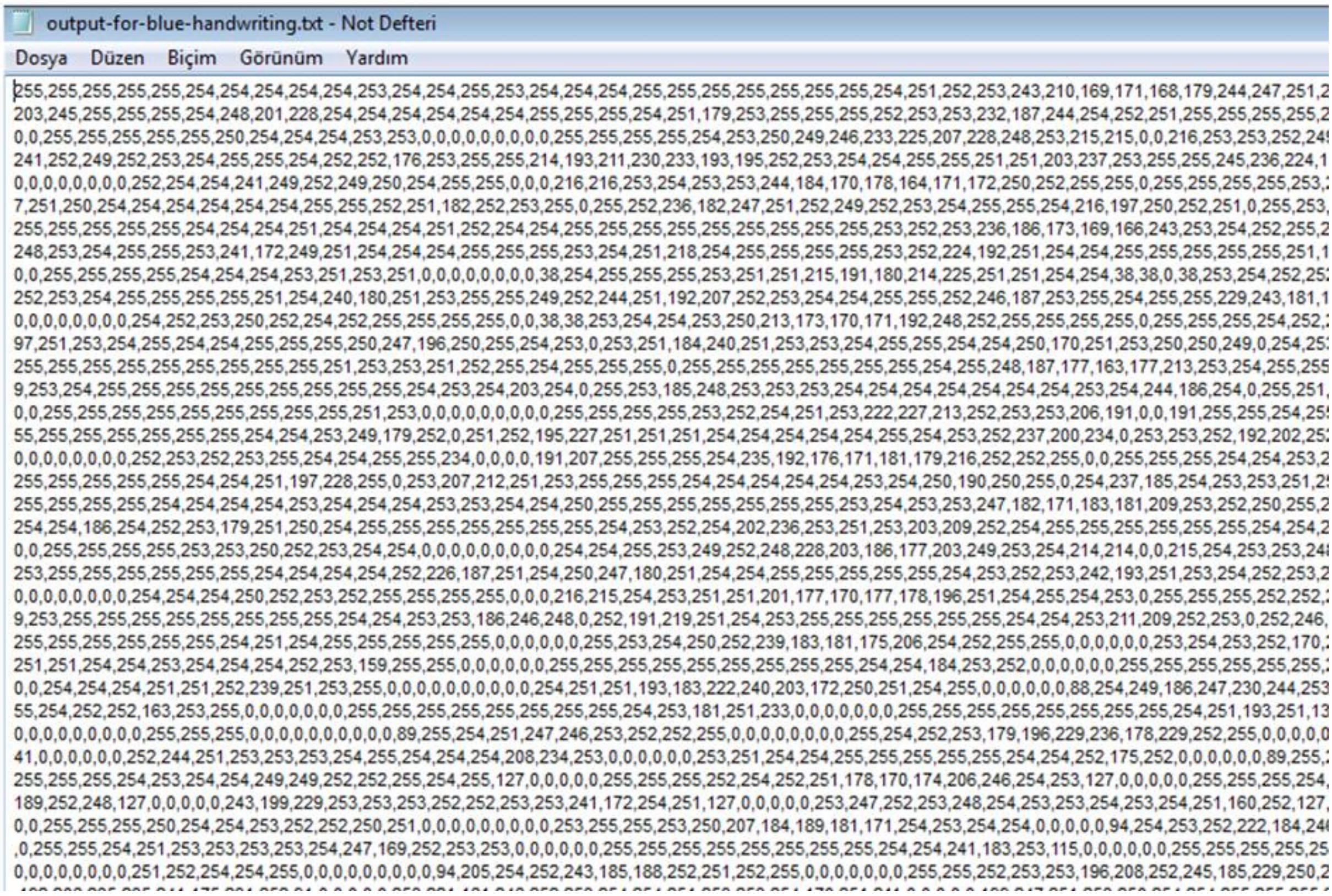

**Figure 4.** Blue colour handwriting image RGB values' file content.

After this training operation, this model's parameters were used for processing test image. In the program this model file was loaded and input image was changed with the test image file. After loading second test image, system creates blobs of each character again. We can again manually change blob size and positions. At the end of all settings, test process will be run with the model file that was generated from training set (alphabetical characters). Each blobs will again have resized to 20x20 pixel images and model file and each blob will be compared. At the end of this process system will give us three possible character name for the blob image. For the OCR process transferring this guessed text to the text classification is last process.

1. Clear and close all visual figures;
2. Clear the terminal screen and move the cursor to the upper left corner.
3. Set input layer size as 400 and number of layers as 26.
4. Load an input text file.
5. Assign "X" and "Y" values from the loaded file content.
6. Calculate "m" as the size of "X".
7. Create a row vector containing a random permutation of 1:m and assign it to "random_indices" variable
8. Select 100 random element of the "X" by using indices located in "random_indices";
9. Display the selected elements.
10. Set "lambda" variable as 0.5
11. Train multiple logistic regression classifiers with the inputs "X", "Y", "num_labels" and "lambda"
12. Returns all and assign the result to "all_theta"
13. Predict the label for a trained one-vs-all classifier using "all_theta" and X.
14. Assign result to "pred" variable
15. Save "all_theta" in an output file as an Ascii content.

**Figure 5.** The pseudo code which is used in Octave the estimate thetas for feature extraction and training.

A screen was created to compare OCR generated text and the real text. After processing test image with the OCR program, generated text was written in a textbox. By clicking compare button we can compare character, word matches and percentages. For handwriting operation accuracy is approximately %60, for computer written images it was %98. For these comparisons “Diff, Match and Patch libraries for Plain Text” was used (https://github.com/google/diff-match-patch). Myer's diff algorithm, which is considered to be the best general purpose diff algorithm, is used by this library. Performance and output quality are improved with the two layers surrounding this algorithm. In addition, the library implements a Bitap matching algorithm.

On the other hand, another method was used which are created by Microsoft. Microsoft Office Document Imaging tool enables OCR operation.

### 3.2 Text Classification with Naive Bayes Algorithm

After OCR process, text files were created which contains OCR generated words. These text files are the test datasets. Training dataset will be the abstracts of journals which are published at Social Science Citation Index by the Institute of Scientific Information. Let’s begin to give some information about Naive Bayes Text Classifier.

Naive Bayes classification algorithm is a classification/categorization algorithm named after Mathematician Thomas Bayes. Naive Bayes classification aims to determine the class, that is, the category of data submitted to the system, with a series of calculations defined according to probability principles. In the Naive Bayes classification, a certain rate of taught data is presented to the system (For example: 1000) The data submitted for teaching must have a class/category. With the probability operations performed on the taught data, the new test data submitted to the system are operated according to the probability values previously obtained, and the category of the given test data is tried to be determined. Of course, the more the number of taught data, the more accurate it can be to identify the true category of test data (Usta, 2014).

## 4. Dataset

For OCR operations, my datasets are scanned images of handwriting and printed documents. You can see the image of these files in Figure 1 and Figure 6.

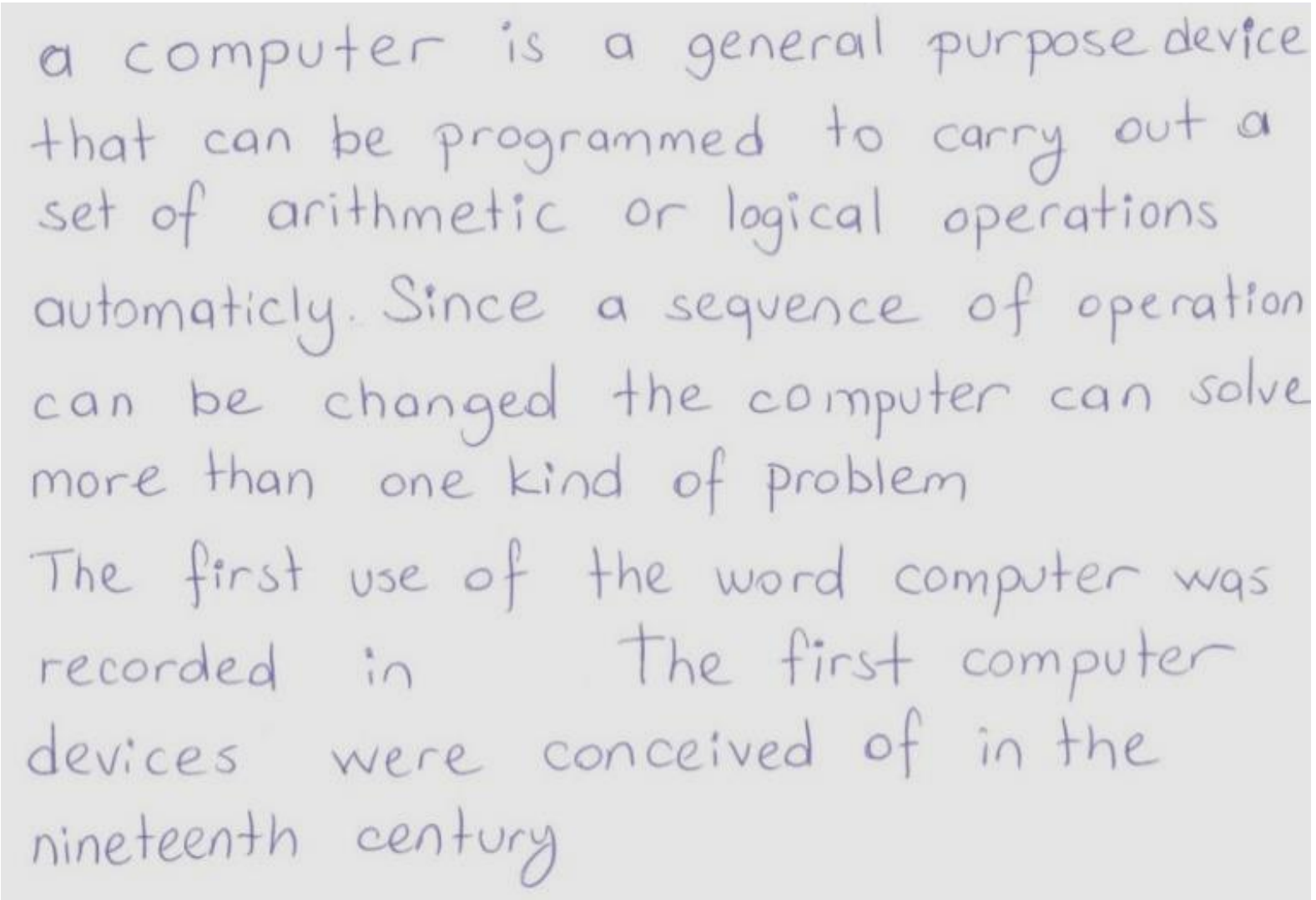
a computer is a general purpose device
that can be programmed to carry out a
set of arithmetic or logical operations
automaticly. Since a sequence of operation
can be changed the computer can solve
more than one kind of problem
The first use of the word computer was
recorded in The first computer
devices were conceived of in the
nineteenth century

Figure 6. Handwriting test dataset for OCR operation.

On the other hand, text classification training set files belongs to two different class. 2200 files were used for training set. This means that approximately 10000 words are processed because five word as significate word was set.

Cisi: Abstracts texts published between 1969 and 1977 and taken from the Social Science Citation Index (SSCI).

Med: Document abstracts in biomedicine received from the National Library of Medicine.

Test files are generated from OCR operation and these abstract files. 295 test files were used with approximately 1000 words.

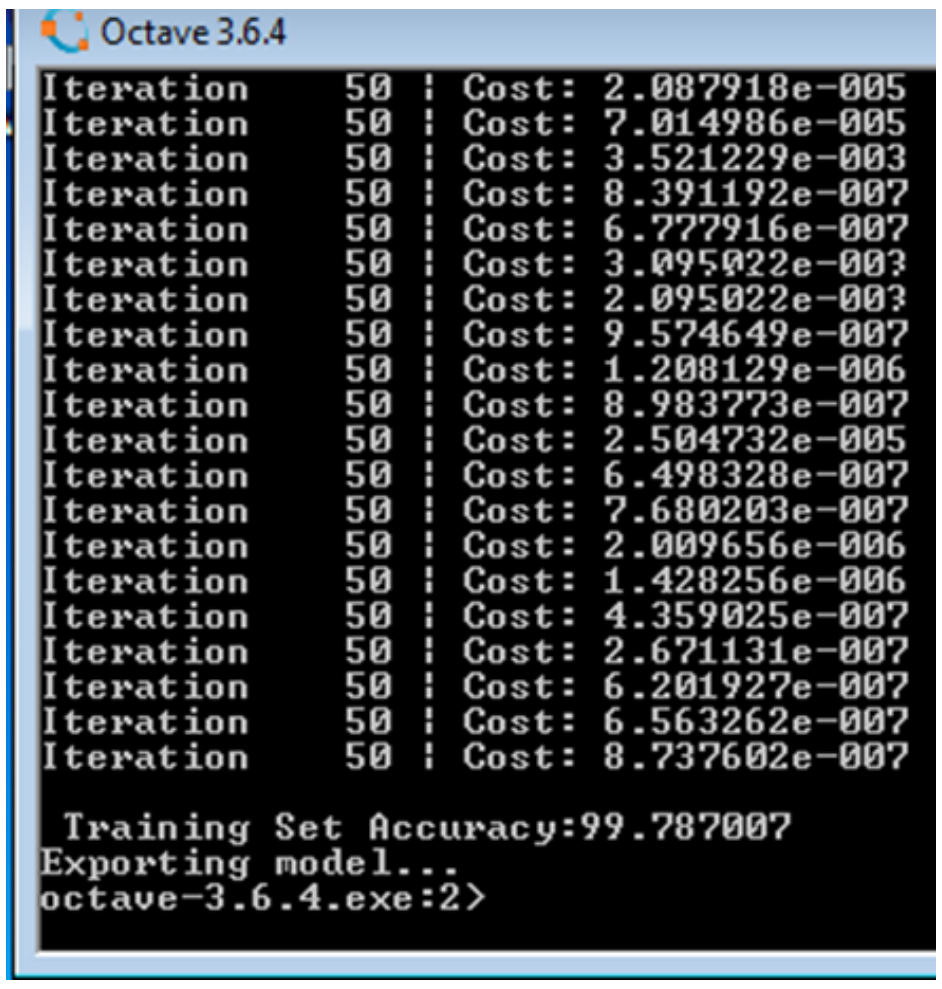

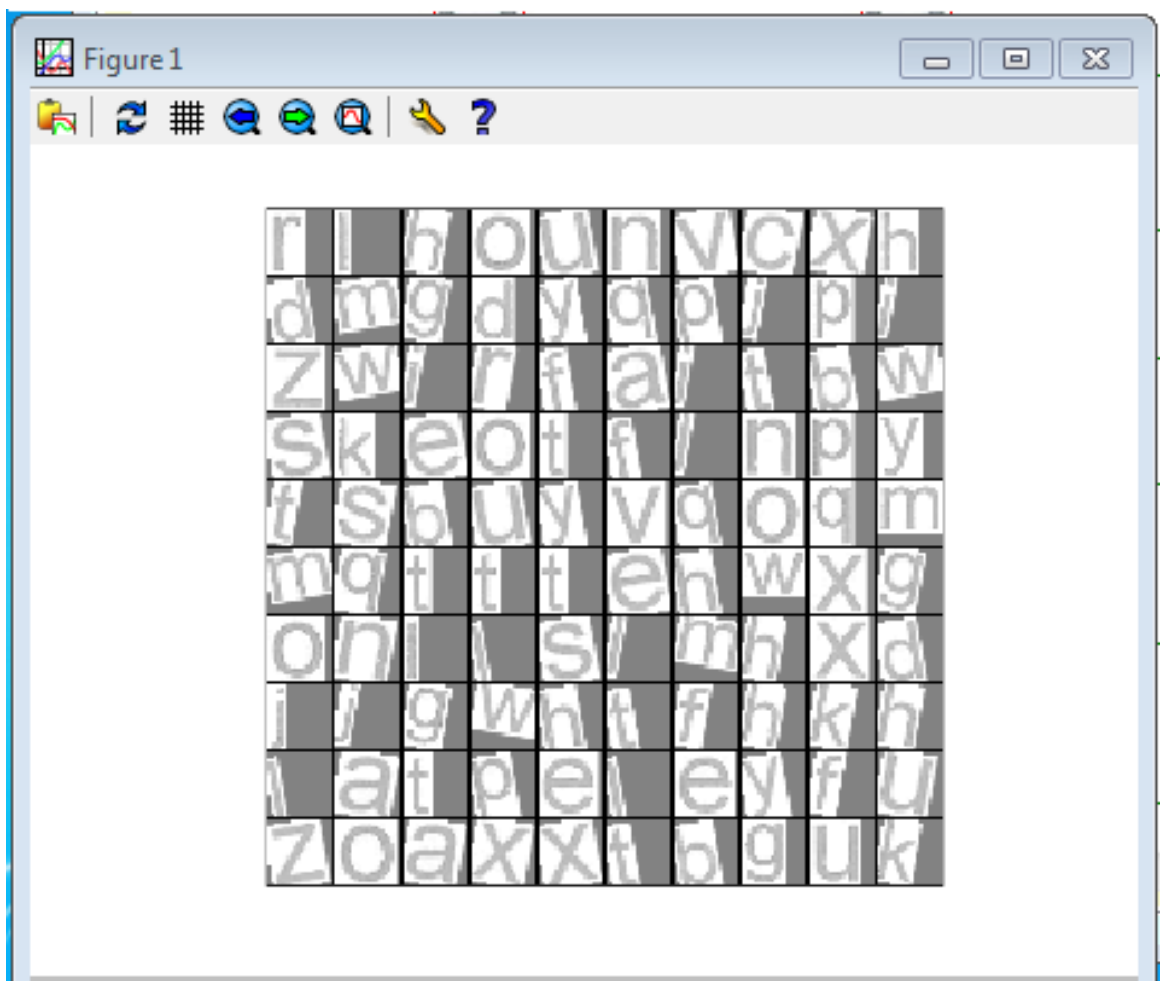

**Figure 7.** Octave Software training operation for OCR.

Text file shown in Figure 4 was given to Octave software for OCR operations and outputs of the OCR training operation are shown in Figure 7.

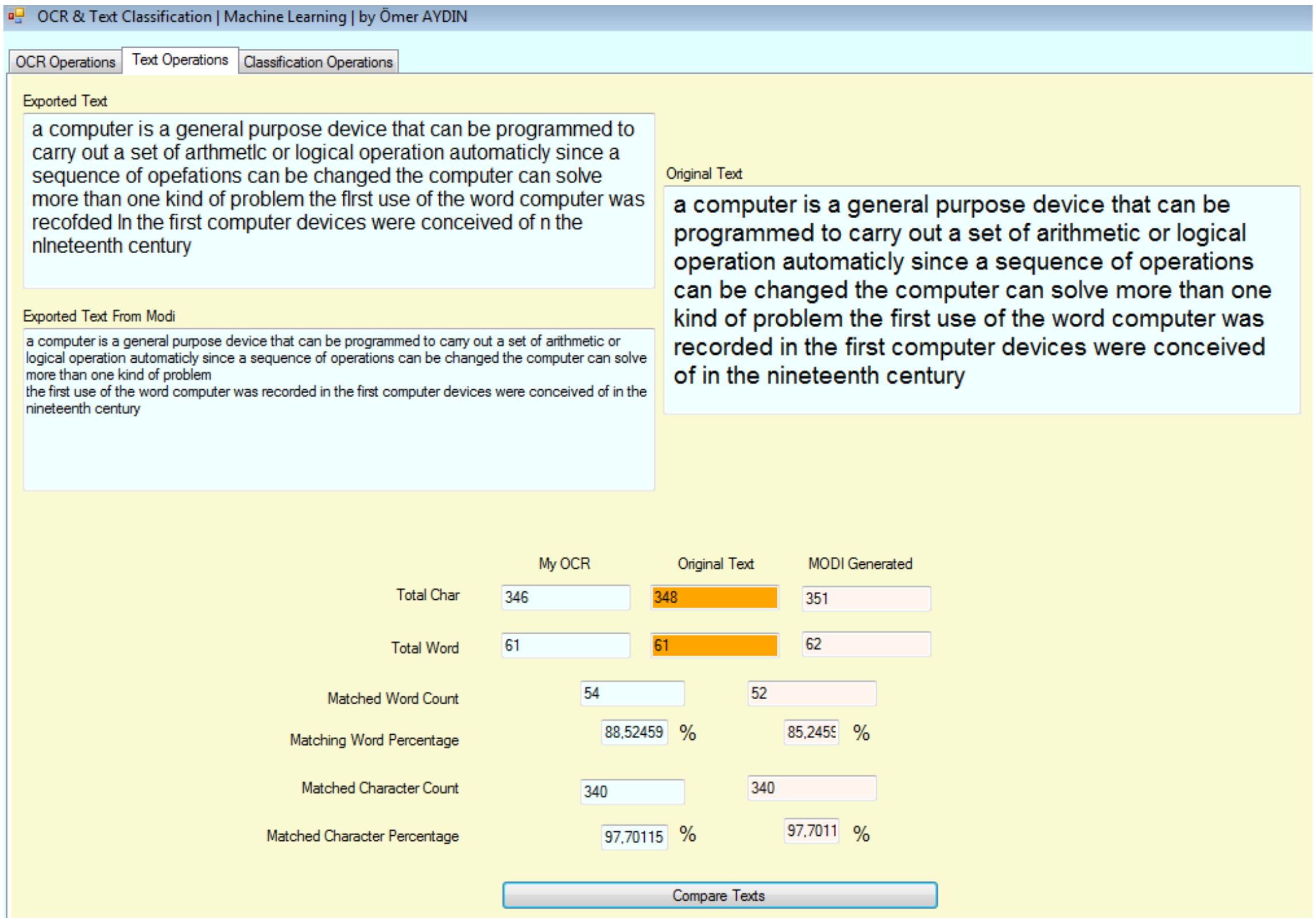

**Figure 8.** Comparison for texts generated by OCR.

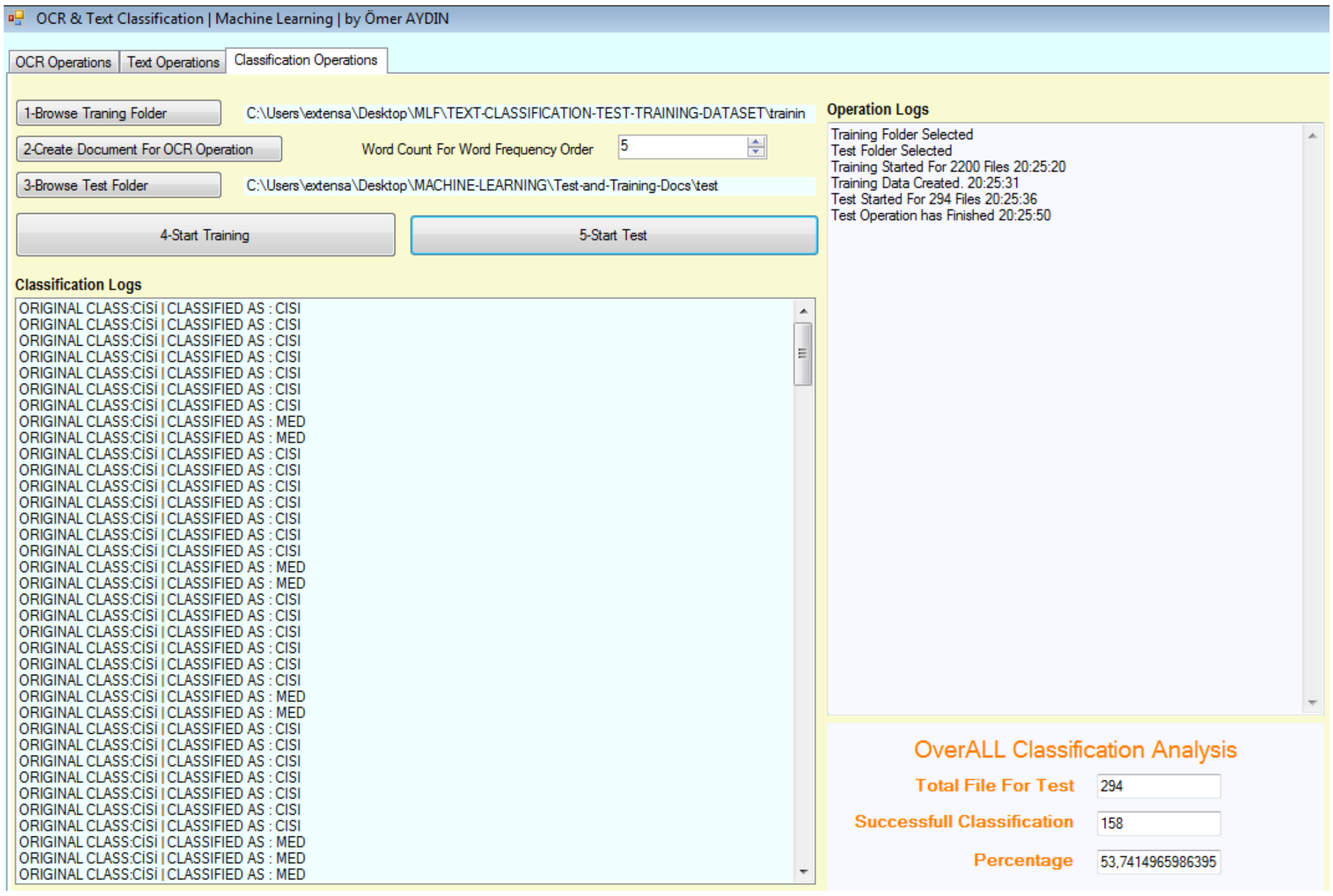

**Figure 9.** Text Classification operations.

## 5. Experimental Results

Two different model were used for extracting text from images. One of them is OCR with Machine Learning, the other is MODI (Microsoft Office Document Imaging Library). The OCR method that is used is getting high accuracy with computer based documents. On the other hand, handwriting documents has very low accuracy. For example, they get scores between %45-%60.

As it can be seen in Figure 8, the original text contains 351 characters and 61 words. According to the data obtained from the OCR system produced with the codes written within the scope of this study, 346 characters and 61 words were determined. On the other hand, 351 characters and 62 words were obtained with MODI. When we look at the match rates of the words, it is seen that the 52 words produced with MODI match the words in the original text. The matching rate was approximately 85.24%. With the OCR codes written within the scope of the study, 54 words were found to match the original text. In this case, a matching ratio of approximately 88.52% was obtained. In terms of character matches, it was seen that 340 characters were matched in both methods, and an approach of 97.7% matching rate was obtained.

As it can be seen in Figure 9, the text classification process was carried out and approximately 53% accuracy was obtained in this classification process using the Naive Bayes method.

## 6. Conclusion

For this project a program was created that is extracting texts from images and classifies this extracted texts. All of these processes are made with Machine Learning methods. The methods use supervised learning. First of all, a stage was created to extract text from images with Optical Character Recognition. Third party GNU software was used to make the training for OCR. After training phase, test dataset images were used to create results. When handwriting documents were used the accuracy of the results were approximately between %45-%60. This result is a bit low. This results can be improved by using different phases. For example, it can be added a dictionary to match the exact word or used a method to get more clear image. You know the quality of the image is very important for OCR process. Also the image can be cleaned and noisy pixels can be removed. For the OCR operation also blobs are very important. If we can create more clear and exact blobs so the result will be better.

For the Naive Bayes phase, the accuracy was %53. We can see better accuracy results in the literature applied by using different methods. Comparing our methods our study is a preliminary study to apply and combine two different methods. This is the main contribution of our study. This accuracy can be improved by selecting more separated training texts. For example, words which are selected for the classification can be unique so this can help to get better results. Also, the training document amount can be more, this will train our algorithm better. Moreover, different classification algorithms and methods such as Neural Network can be used in future works.